%% file: xlmr.tex
\documentclass[11pt,a4paper]{article}
\usepackage[hyperref]{acl2020}
\usepackage{times}
\usepackage{latexsym}

\usepackage{microtype}
\usepackage{graphicx}
\usepackage{subfigure}
\usepackage{booktabs} 
\usepackage{url}
\usepackage{times}
\usepackage{latexsym}
\usepackage{array}
\usepackage{adjustbox}
\usepackage{multirow}
\usepackage{hyperref}
\usepackage{longtable}

\input{content/tables}

\aclfinalcopy 
\def\aclpaperid{479} 

\usepackage{xspace}
\newcommand{\xlmr}{\textit{XLM-R}\xspace}
\newcommand{\mbert}{mBERT\xspace}

\title{Unsupervised Cross-lingual Representation Learning at Scale}

\author{Alexis Conneau\thanks{\ \ Equal contribution.} \space\space\space
  Kartikay Khandelwal\footnotemark[1] \space\space\space \AND
  \bf Naman Goyal \space\space\space 
  Vishrav Chaudhary \space\space\space
  Guillaume Wenzek \space\space\space 
  Francisco Guzm\'an \space\space\space \AND
  \bf Edouard Grave \space\space\space
  Myle Ott \space\space\space
  Luke Zettlemoyer \space\space\space
  Veselin Stoyanov \space\space\space \\ \\ \\
  \bf Facebook AI
  }

\date{}

\begin{document}
\maketitle
\begin{abstract}
This paper shows that pretraining  multilingual language models at scale leads to significant performance gains for a wide range of cross-lingual transfer tasks. We train a Transformer-based masked language model on one hundred languages, using more than two terabytes of filtered CommonCrawl data. Our model, dubbed \xlmr, significantly outperforms multilingual BERT (mBERT) on a variety of cross-lingual benchmarks, including +14.6\% average accuracy on XNLI, +13\% average F1 score on MLQA, and +2.4\% F1 score on NER. \xlmr performs particularly well on low-resource languages, improving 15.7\% in XNLI accuracy for Swahili and 11.4\% for Urdu over previous XLM models. We also present a detailed empirical analysis of the key factors that are required to achieve these gains, including the trade-offs between (1) positive transfer and capacity dilution and (2) the performance of high and low resource languages at scale. Finally, we show, for the first time, the possibility of multilingual modeling without sacrificing per-language performance; \xlmr is very competitive with strong monolingual models on the GLUE and XNLI benchmarks. We will make our code, data and models publicly available.{\let\thefootnote\relax\footnotetext{\scriptsize Correspondence to {\tt \{aconneau,kartikayk\}@fb.com}}}\footnote{\url{https://github.com/facebookresearch/(fairseq-py,pytext,xlm)}}
\end{abstract}

\section{Introduction}

The goal of this paper is to improve cross-lingual language understanding (XLU), by carefully studying the effects of training unsupervised cross-lingual representations at a very large scale. 
We present \xlmr a transformer-based multilingual masked language model pre-trained on text in 100 languages, which obtains state-of-the-art performance on cross-lingual classification, sequence labeling and question answering.

Multilingual masked language models (MLM) like \mbert ~\cite{devlin2018bert} and XLM \cite{lample2019cross} have pushed the state-of-the-art on cross-lingual understanding tasks by jointly pretraining large Transformer models~\cite{transformer17} on many languages. These models allow for effective cross-lingual transfer, as seen in a number of benchmarks including cross-lingual natural language inference ~\cite{bowman2015large,williams2017broad,conneau2018xnli}, question answering~\cite{rajpurkar-etal-2016-squad,lewis2019mlqa}, and named entity recognition~\cite{Pires2019HowMI,wu2019beto}.
However, all of these studies pre-train on Wikipedia, which provides a relatively limited scale especially for lower resource languages.

In this paper, we first present a comprehensive analysis of the trade-offs and limitations of multilingual language models at scale, inspired by recent monolingual scaling efforts~\cite{roberta2019}.
We measure the trade-off between high-resource and low-resource languages and the impact of language sampling and vocabulary size. 
The experiments expose a trade-off as we scale the number of languages for a fixed model capacity: more languages leads to better cross-lingual performance on low-resource languages up until a point, after which the overall performance on monolingual and cross-lingual benchmarks degrades. We refer to this tradeoff as the \emph{curse of multilinguality}, and show that it can be alleviated by simply increasing model capacity.
We argue, however, that this remains an important limitation for future XLU systems which may aim to improve performance with more modest computational budgets. 

Our best model XLM-RoBERTa (\xlmr)   outperforms \mbert on cross-lingual classification by up to 23\% accuracy on low-resource languages.
It outperforms the previous state of the art by 5.1\% average accuracy on XNLI, 2.42\% average F1-score on Named Entity Recognition, and 9.1\% average F1-score on cross-lingual Question Answering. We also evaluate monolingual fine tuning on the GLUE and XNLI benchmarks, where \xlmr obtains results competitive with state-of-the-art monolingual models, including RoBERTa \cite{roberta2019}. 
These results demonstrate, for the first time, that it is possible to have a single large model for all languages, without sacrificing per-language performance.
We will make our code, models and data publicly available, with the hope that this will help research in multilingual NLP and low-resource language understanding.

\section{Related Work}
From pretrained word embeddings~\citep{mikolov2013distributed, pennington2014glove} to pretrained contextualized representations~\citep{peters2018deep,schuster2019cross} and transformer based language models~\citep{radford2018improving,devlin2018bert}, unsupervised representation learning has significantly improved the state of the art in natural language understanding. Parallel work on cross-lingual understanding~\citep{mikolov2013exploiting,schuster2019cross,lample2019cross} extends these systems to more languages and to the cross-lingual setting in which a model is learned in one language and applied in other languages. 

Most recently, \citet{devlin2018bert} and \citet{lample2019cross} introduced \mbert and XLM - masked language models trained on multiple languages, without any cross-lingual supervision. 
\citet{lample2019cross} propose translation language modeling (TLM) as a way to leverage parallel data and obtain a new state of the art on the cross-lingual natural language inference (XNLI) benchmark~\cite{conneau2018xnli}. 
They further show strong improvements on unsupervised machine translation and pretraining for sequence generation. \citet{wu2019emerging} shows that monolingual BERT representations are similar across languages, explaining in part the natural emergence of multilinguality in bottleneck architectures. Separately, \citet{Pires2019HowMI} demonstrated the effectiveness of multilingual models like \mbert on sequence labeling tasks. \citet{huang2019unicoder} showed gains over XLM using cross-lingual multi-task learning, and \citet{singh2019xlda} demonstrated the efficiency of cross-lingual data augmentation for cross-lingual NLI. However, all of this work was at a relatively modest scale, in terms of the amount of training data, as compared to our approach.

\insertWikivsCC

The benefits of scaling language model pretraining by increasing the size of the model as well as the training data has been extensively studied in the literature. For the monolingual case, \citet{jozefowicz2016exploring} show how large-scale LSTM models can obtain much stronger performance on language modeling benchmarks when trained on billions of tokens.
GPT~\cite{radford2018improving} also highlights the importance of scaling the amount of data and RoBERTa \cite{roberta2019} shows that training BERT longer on more data leads to significant boost in performance. Inspired by RoBERTa, we show that mBERT and XLM are undertuned, and that simple improvements in the learning procedure of unsupervised MLM leads to much better performance. We train on cleaned CommonCrawls~\cite{wenzek2019ccnet}, which increase the amount of data for low-resource languages by two orders of magnitude on average. Similar data has also been shown to be effective for learning high quality word embeddings in multiple languages~\cite{grave2018learning}.

Several efforts have trained massively multilingual machine translation models from large parallel corpora. They uncover the high and low resource trade-off and the problem of capacity dilution~\citep{johnson2017google,tan2019multilingual}. The work most similar to ours is \citet{arivazhagan2019massively}, which trains a single model in 103 languages on over 25 billion parallel sentences. 
\citet{siddhant2019evaluating} further analyze the representations obtained by the encoder of a massively multilingual machine translation system and show that it obtains similar results to mBERT on cross-lingual NLI.
Our work, in contrast, focuses on the unsupervised learning of cross-lingual representations and their transfer to discriminative tasks.

\section{Model and Data}
\label{sec:model+data}

In this section, we present the training objective, languages, and data we use. We follow the XLM approach~\cite{lample2019cross} as closely as possible, only introducing changes that improve performance at scale.  

\paragraph{Masked Language Models.}
We use a Transformer model~\cite{transformer17} trained with the multilingual MLM objective~\cite{devlin2018bert,lample2019cross} using only monolingual data. We sample streams of text from each language and train the model to predict the masked tokens in the input. 
We apply subword tokenization directly on raw text data using Sentence Piece~\cite{kudo2018sentencepiece} with a unigram language model~\cite{kudo2018subword}. We sample batches from different languages using the same sampling distribution as \citet{lample2019cross}, but with $\alpha=0.3$. Unlike \citet{lample2019cross}, we do not use language embeddings, which allows our model to better deal with code-switching.  We use a large vocabulary size of 250K with a full softmax and train two different models: \xlmr\textsubscript{Base} (L = 12, H = 768, A = 12, 270M params) and \xlmr (L = 24, H = 1024, A = 16, 550M params). For all of our ablation studies, we use a BERT\textsubscript{Base} architecture with a vocabulary of 150K tokens. Appendix~\ref{sec:appendix_B} goes into more details about the architecture of the different models referenced in this paper. 

\paragraph{Scaling to a hundred languages.}
\xlmr is trained on 100 languages;
we provide a full list of languages and associated statistics in Appendix~\ref{sec:appendix_A}. Figure~\ref{fig:wikivscc} specifies the iso codes of 88 languages that are shared across \xlmr and XLM-100, the model from~\citet{lample2019cross} trained on Wikipedia text in 100 languages. 

Compared to previous work, we replace some languages with more commonly used ones such as romanized Hindi and traditional Chinese. In our ablation studies, we always include the 7 languages for which we have classification and sequence labeling evaluation benchmarks: English, French, German, Russian, Chinese, Swahili and Urdu. We chose this set as it covers a suitable range of language families and includes low-resource languages such as Swahili and Urdu.
We also consider larger sets of 15, 30, 60 and all 100 languages. When reporting results on high-resource and low-resource, we refer to the average of English and French results, and the average of Swahili and Urdu results respectively.

\paragraph{Scaling the Amount of Training Data.}
Following~\citet{wenzek2019ccnet}~\footnote{\url{https://github.com/facebookresearch/cc_net}}, we build a clean CommonCrawl Corpus in 100 languages. We use an internal language identification model in combination with the one from fastText~\cite{joulin2017bag}. We train language models in each language and use it to filter documents as described in \citet{wenzek2019ccnet}. We consider one CommonCrawl dump for English and twelve dumps for all other languages, which significantly increases dataset sizes, especially for low-resource languages like Burmese and Swahili. 

Figure~\ref{fig:wikivscc} shows the difference in size between the Wikipedia Corpus used by mBERT and XLM-100, and the CommonCrawl Corpus we use. As we show in Section~\ref{sec:multimono}, monolingual Wikipedia corpora are too small to enable unsupervised representation learning. Based on our experiments, we found that a few hundred MiB of text data is usually a minimal size for learning a BERT model.

\section{Evaluation}
We consider four evaluation benchmarks. 
For cross-lingual understanding, we use cross-lingual natural language inference, named entity recognition, and question answering. We use the GLUE benchmark to evaluate the English performance of \xlmr and compare it to other state-of-the-art models.

\paragraph{Cross-lingual Natural Language Inference (XNLI).}
The XNLI dataset comes with ground-truth dev and test sets in 15 languages, and a ground-truth English training set. The training set has been machine-translated to the remaining 14 languages, providing synthetic training data for these languages as well. We evaluate our model on cross-lingual transfer from English to other languages. We also consider three machine translation baselines: (i) \textit{translate-test}: dev and test sets are machine-translated to English and a single English model is used (ii) \textit{translate-train} (per-language): the English training set is machine-translated to each language and we fine-tune a multiligual model on each training set (iii) \textit{translate-train-all} (multi-language): we fine-tune a multilingual model on the concatenation of all training sets from translate-train. For the translations, we use the official data provided by the XNLI project.

\paragraph{Named Entity Recognition.}
For NER, we consider the CoNLL-2002~\cite{sang2002introduction} and CoNLL-2003~\cite{tjong2003introduction} datasets in English, Dutch, Spanish and German. We fine-tune multilingual models either (1) on the English set to evaluate cross-lingual transfer, (2) on each set to  evaluate per-language performance, or (3) on all sets to evaluate multilingual learning. We report the F1 score, and compare to baselines from \citet{lample-etal-2016-neural} and \citet{akbik2018coling}.

\paragraph{Cross-lingual Question Answering.}
We use the MLQA benchmark from \citet{lewis2019mlqa}, which extends the English SQuAD benchmark to Spanish, German, Arabic, Hindi, Vietnamese and Chinese. We report the F1 score as well as the exact match (EM) score for cross-lingual transfer from English.

\paragraph{GLUE Benchmark.}
Finally, we evaluate the English performance of our model on the GLUE benchmark~\cite{wang2018glue} which gathers multiple classification tasks, such as MNLI~\cite{williams2017broad}, SST-2~\cite{socher2013recursive}, or QNLI~\cite{rajpurkar2018know}. We use BERT\textsubscript{Large} and RoBERTa as baselines.

\section{Analysis and Results}
\label{sec:analysis}

In this section, we perform a comprehensive analysis of multilingual masked language models. We conduct most of the analysis on XNLI, which we found to be representative of our findings on other tasks. We then present the results of \xlmr on cross-lingual understanding and GLUE. Finally, we compare multilingual and monolingual models, and present results on low-resource languages.

\subsection{Improving and Understanding Multilingual Masked Language Models}
\insertAblationone
\insertAblationtwo

Much of the work done on understanding the cross-lingual effectiveness of \mbert or XLM~\cite{Pires2019HowMI,wu2019beto,lewis2019mlqa} has focused on analyzing the performance of fixed pretrained models on downstream tasks. In this section, we present a comprehensive study of different factors that are important to \textit{pretraining} large scale multilingual models. We highlight the trade-offs and limitations of these models as we scale to one hundred languages.

\paragraph{Transfer-dilution Trade-off and Curse of Multilinguality.}
Model capacity (i.e. the number of parameters in the model) is constrained due to practical considerations such as memory and speed during training and inference. For a fixed sized model, the per-language capacity decreases as we increase the number of languages. While low-resource language performance can be improved by adding similar higher-resource languages during pretraining, the overall downstream performance suffers from this capacity dilution~\cite{arivazhagan2019massively}. Positive transfer and capacity dilution have to be traded off against each other. 

We illustrate this trade-off in Figure~\ref{fig:transfer_dilution}, which shows XNLI performance vs the number of languages the model is pretrained on. Initially, as we go from 7 to 15 languages, the model is able to take advantage of positive transfer which improves performance, especially on low resource languages. Beyond this point the {\em curse of multilinguality}
kicks in and degrades performance across all languages.  Specifically, the overall XNLI accuracy decreases from 71.8\% to 67.7\% as we go from XLM-7 to XLM-100. The same trend can be observed for models trained on the larger CommonCrawl Corpus. 

The issue is even more prominent when the capacity of the model is small. To show this, we pretrain models on Wikipedia Data in 7, 30 and 100 languages. As we add more languages, we make the Transformer wider by increasing the hidden size from 768 to 960 to 1152. In Figure~\ref{fig:capacity}, we show that the added capacity allows XLM-30 to be on par with XLM-7, thus overcoming the curse of multilinguality. The added capacity for XLM-100, however, is not enough 
and it still lags behind due to higher vocabulary dilution (recall from Section~\ref{sec:model+data} that we used a fixed vocabulary size of 150K for all models). 

\paragraph{High-resource vs Low-resource Trade-off.}
The allocation of the model capacity across languages is controlled by several parameters: the training set size, the size of the shared subword vocabulary, and the rate at which we sample training examples from each language. We study the effect of sampling on the performance of high-resource (English and French) and low-resource (Swahili and Urdu) languages for an XLM-100 model trained on Wikipedia (we observe a similar trend for the construction of the subword vocab). Specifically, we investigate the impact of varying the $\alpha$ parameter which controls the exponential smoothing of the language sampling rate. Similar to~\citet{lample2019cross}, we use a sampling rate proportional to the number of sentences in each corpus.  Models trained with higher values of $\alpha$ see batches of high-resource languages more often. 
Figure~\ref{fig:alpha} shows that the higher the value of $\alpha$, the better the performance on high-resource languages, and vice-versa. When considering overall performance, we found $0.3$ to be an optimal value for $\alpha$, and use this for \xlmr.

\paragraph{Importance of Capacity and Vocabulary.}
In previous sections and in Figure~\ref{fig:capacity}, we showed the importance of scaling the model size as we increase the number of languages. Similar to the overall model size, we argue that scaling the size of the shared vocabulary (the vocabulary capacity) can improve the performance of multilingual models on downstream tasks. To illustrate this effect, we train XLM-100 models on Wikipedia data with different vocabulary sizes. We keep the overall number of parameters constant by adjusting the width of the transformer. Figure~\ref{fig:vocab} shows that even with a fixed capacity, we observe a 2.8\% increase in XNLI average accuracy as we increase the vocabulary size from 32K to 256K. This suggests that multilingual models can benefit from allocating a higher proportion of the total number of parameters to the embedding layer even though this reduces the size of the Transformer.
For simplicity and given the softmax computational constraints, we use a vocabulary of 250k for \xlmr.

We further illustrate the importance of this parameter, by training three models with the same transformer architecture (BERT\textsubscript{Base}) but with different vocabulary sizes: 128K, 256K and 512K. We observe more than 3\% gains in overall accuracy on XNLI by simply increasing the vocab size from 128k to 512k.

\paragraph{Larger-scale Datasets and Training.}
As shown in Figure~\ref{fig:wikivscc}, the CommonCrawl Corpus that we collected has significantly more monolingual data than the previously used Wikipedia corpora. Figure~\ref{fig:curse} shows that for the same BERT\textsubscript{Base} architecture, all models trained on CommonCrawl obtain significantly better performance.

Apart from scaling the training data, \citet{roberta2019} also showed the benefits of training MLMs longer. In our experiments, we observed similar effects of large-scale training, such as increasing batch size (see Figure~\ref{fig:batch}) and training time, on model performance. Specifically, we found that using validation perplexity as a stopping criterion for pretraining caused the multilingual MLM in \citet{lample2019cross} to be under-tuned. In our experience, performance on downstream tasks continues to improve even after validation perplexity has plateaued. Combining this observation with our implementation of the unsupervised XLM-MLM objective, we were able to improve the performance of \citet{lample2019cross} from 71.3\% to more than 75\% average accuracy on XNLI, which was on par with their supervised translation language modeling (TLM) objective. Based on these results, and given our focus on unsupervised learning, we decided to not use the supervised TLM objective for training our models.

\paragraph{Simplifying Multilingual Tokenization with Sentence Piece.}
The different language-specific tokenization tools 
used by mBERT and XLM-100 make these models more difficult to use on raw text. Instead, we train a Sentence Piece model (SPM) and apply it directly on raw text data for all languages. We did not observe any loss in performance for models trained with SPM when compared to models trained with language-specific preprocessing and byte-pair encoding (see Figure~\ref{fig:batch}) and hence use SPM for \xlmr.

\subsection{Cross-lingual Understanding Results}
Based on these results, we adapt the setting of \citet{lample2019cross} and use a large Transformer model with 24 layers and 1024 hidden states, with a 250k vocabulary. We use the multilingual MLM loss and train our \xlmr model for 1.5 Million updates on five-hundred 32GB Nvidia V100 GPUs with a batch size of 8192. We leverage the SPM-preprocessed text data from CommonCrawl in 100 languages and sample languages with $\alpha=0.3$. In this section, we show that it outperforms all previous techniques on cross-lingual benchmarks while getting performance on par with RoBERTa on the GLUE benchmark.

\insertXNLItable

\paragraph{XNLI.}
Table~\ref{tab:xnli} shows XNLI results and adds some additional details: (i) the number of models the approach induces (\#M), (ii) the data on which the model was trained (D), and (iii) the number of languages the model was pretrained on (\#lg). As we show in our results, these parameters significantly impact performance. Column \#M specifies whether model selection was done separately on the dev set of each language ($N$ models), or on the joint dev set of all the languages (single model). We observe a 0.6 decrease in overall accuracy when we go from $N$ models to a single model - going from 71.3 to 70.7. We encourage the community to adopt this setting. For cross-lingual transfer, while this approach is not fully zero-shot transfer, we argue that in real applications, a small amount of supervised data is often available for validation in each language.

\xlmr sets a new state of the art on XNLI.
On cross-lingual transfer, \xlmr obtains 80.9\%  accuracy, outperforming the XLM-100 and \mbert open-source models by 10.2\% and 14.6\% average accuracy. On the Swahili and Urdu low-resource languages, \xlmr outperforms XLM-100 by 15.7\% and 11.4\%, and \mbert by 23.5\% and 15.8\%. While \xlmr handles 100 languages, we also show that it outperforms the former state of the art Unicoder~\citep{huang2019unicoder} and XLM (MLM+TLM),  which handle only 15 languages, by 5.5\% and 5.8\% average accuracy respectively. Using the multilingual training of translate-train-all, \xlmr further improves performance and reaches 83.6\% accuracy, a new overall state of the art for XNLI, outperforming Unicoder by 5.1\%. Multilingual training is similar to practical applications where training sets are available in various languages for the same task. In the case of XNLI, datasets have been translated, and translate-train-all can be seen as some form of cross-lingual data augmentation~\cite{singh2019xlda}, similar to back-translation~\cite{xie2019unsupervised}.

\insertNER
\paragraph{Named Entity Recognition.}
In Table~\ref{tab:ner}, we report results of \xlmr and \mbert on CoNLL-2002 and CoNLL-2003. We consider the LSTM + CRF approach from \citet{lample-etal-2016-neural} and the Flair model from \citet{akbik2018coling} as baselines. We evaluate the performance of the model on each of the target languages in three different settings: (i) train on English data only (en) (ii) train on data in target language (each) (iii) train on data in all languages (all). Results of \mbert are reported from \citet{wu2019beto}. Note that we do not use a linear-chain CRF on top of \xlmr and \mbert representations, which gives an advantage to \citet{akbik2018coling}. Without the CRF, our \xlmr model still performs on par with the state of the art, outperforming \citet{akbik2018coling} on Dutch by $2.09$ points. On this task, \xlmr also outperforms \mbert by 2.42 F1 on average for cross-lingual transfer, and 1.86 F1 when trained on each language. Training on all languages leads to an average F1 score of 89.43\%, outperforming cross-lingual transfer approach by 8.49\%.

\paragraph{Question Answering.}
We also obtain new state of the art results on the MLQA cross-lingual question answering benchmark, introduced by \citet{lewis2019mlqa}. We follow their procedure by training on the English training data and evaluating on the 7 languages of the dataset.
We report results in Table~\ref{tab:mlqa}. 
\xlmr obtains F1 and accuracy scores of 70.7\% and 52.7\% while the previous state of the art was 61.6\% and 43.5\%. \xlmr also outperforms \mbert by 13.0\% F1-score and 11.1\% accuracy. It even outperforms BERT-Large on English, confirming its strong monolingual performance.

\insertMLQA

\subsection{Multilingual versus Monolingual}
\label{sec:multimono}
In this section, we present results of multilingual XLM models against monolingual BERT models.

\paragraph{GLUE: \xlmr versus RoBERTa.}
Our goal is to obtain a multilingual model with strong performance on both, cross-lingual understanding tasks as well as natural language understanding tasks for each language. To that end, we evaluate \xlmr on the GLUE benchmark. We show in Table~\ref{tab:glue}, that \xlmr obtains better average dev performance than BERT\textsubscript{Large} by 1.6\% and reaches performance on par with XLNet\textsubscript{Large}. The RoBERTa model outperforms \xlmr by only 1.0\% on average. We believe future work can reduce this gap even further by alleviating the curse of multilinguality and vocabulary dilution. These results demonstrate the possibility of learning one model for many languages while maintaining strong performance on per-language downstream tasks.

\insertGlue

\paragraph{XNLI: XLM versus BERT.}
A recurrent criticism against multilingual models is that they obtain worse performance than their monolingual counterparts. In addition to the comparison of \xlmr and RoBERTa, we provide the first comprehensive study to assess this claim on the XNLI benchmark. We extend our comparison between multilingual XLM models and monolingual BERT models on 7 languages and compare performance in Table~\ref{tab:multimono}. We train 14 monolingual BERT models on Wikipedia and CommonCrawl (capped at 60 GiB),
and two XLM-7 models. We increase the vocabulary size of the multilingual model for a better comparison.
We found that \textit{multilingual models can outperform their monolingual BERT counterparts}. Specifically, in Table~\ref{tab:multimono}, we show that for cross-lingual transfer, monolingual baselines outperform XLM-7 for both Wikipedia and CC by 1.6\% and 1.3\% average accuracy. However, by making use of multilingual training (translate-train-all) and leveraging training sets coming from multiple languages, XLM-7 can outperform the BERT models: our XLM-7 trained on CC obtains 80.0\% average accuracy on the 7 languages, while the average performance of BERT models trained on CC is 77.5\%. This is a surprising result that shows that the capacity of multilingual models to leverage training data coming from multiple languages for a particular task can overcome the capacity dilution problem to obtain better overall performance.

\insertMultiMono

\subsection{Representation Learning for Low-resource Languages}
We observed in Table~\ref{tab:multimono} that pretraining on Wikipedia for Swahili and Urdu performed similarly to a randomly initialized model; most likely due to the small size of the data for these languages. On the other hand, pretraining on CC improved performance by up to 10 points. This confirms our assumption that mBERT and XLM-100 rely heavily on cross-lingual transfer but do not model the low-resource languages as well as \xlmr. Specifically, in the translate-train-all setting, we observe that the biggest gains for XLM models trained on CC, compared to their Wikipedia counterparts, are on low-resource languages; 7\% and 4.8\% improvement on Swahili and Urdu respectively.

\section{Conclusion}
In this work, we introduced \xlmr, our new state of the art multilingual masked language model trained on 2.5 TB of newly created clean CommonCrawl data in 100 languages. We show that it provides strong gains over previous multilingual models like \mbert and XLM on classification, sequence labeling and question answering. We exposed the limitations of multilingual MLMs, in particular by uncovering the high-resource versus low-resource trade-off, the curse of multilinguality and the importance of key hyperparameters. We also expose the surprising effectiveness of multilingual models over monolingual models, and show strong improvements on low-resource languages.

\bibliography{acl2020}
\bibliographystyle{acl_natbib}

 \newpage
 \clearpage
 \appendix
 \onecolumn
 \section*{Appendix}
 \section{Languages and statistics for CC-100 used by \xlmr}
 In this section we present the list of languages in the CC-100 corpus we created for training \xlmr. We also report statistics such as the number of tokens and the size of each monolingual corpus.
 \label{sec:appendix_A}
 \insertDataStatistics

 \section{Model Architectures and Sizes}
 As we showed in section~\ref{sec:analysis}, capacity is an important parameter for learning strong cross-lingual representations. In the table below, we list multiple monolingual and multilingual models used by the research community and summarize their architectures and total number of parameters.
 \label{sec:appendix_B}

\insertParameters





\end{document}


\nobibliography{acl2020}
\bibliographystyle{acl_natbib}
\appendix
\onecolumn
\section*{Supplementary materials}
\section{Languages and statistics for CC-100 used by \xlmr}
In this section we present the list of languages in the CC-100 corpus we created for training \xlmr. We also report statistics such as the number of tokens and the size of each monolingual corpus.
\label{sec:appendix_A}
\insertDataStatistics

\newpage
\section{Model Architectures and Sizes}
As we showed in section 5, capacity is an important parameter for learning strong cross-lingual representations. In the table below, we list multiple monolingual and multilingual models used by the research community and summarize their architectures and total number of parameters.
\label{sec:appendix_B}

\insertParameters

%% file: content/tables.tex
\newcommand{\insertXNLItable}{
    \begin{table*}[h!]
        \begin{center}
            \resizebox{1\linewidth}{!}{
            \begin{tabular}[b]{l ccc ccccccccccccccc c}
            \toprule
            {\bf Model} & {\bf D }& {\bf \#M} & {\bf \#lg} & {\bf en} & {\bf fr} & {\bf es} & {\bf de} & {\bf el} & {\bf bg} & {\bf ru} & {\bf tr} & {\bf ar} & {\bf vi} & {\bf th} & {\bf zh} & {\bf hi} & {\bf sw} & {\bf ur} & {\bf Avg}\\
                \midrule
                
                \multicolumn{19}{l}{\it Fine-tune multilingual model on English training set (Cross-lingual Transfer)} \\
                \midrule
                \citet{lample2019cross} & Wiki+MT & N & 15 & 85.0 & 78.7 & 78.9 & 77.8 & 76.6 & 77.4 & 75.3 & 72.5 & 73.1 & 76.1 & 73.2 & 76.5 & 69.6 & 68.4 & 67.3 & 75.1 \\
                \citet{huang2019unicoder} & Wiki+MT & N & 15 & 85.1 & 79.0 & 79.4 & 77.8 & 77.2 & 77.2 & 76.3 & 72.8 & 73.5 & 76.4 & 73.6 & 76.2 & 69.4 & 69.7 & 66.7 & 75.4 \\
                \citet{devlin2018bert} & Wiki & N & 102 & 82.1 & 73.8 & 74.3 & 71.1 & 66.4 & 68.9 & 69.0 & 61.6 & 64.9 & 69.5 & 55.8 & 69.3 & 60.0 & 50.4 & 58.0 & 66.3  \\
                \citet{lample2019cross}  & Wiki & N & 100 & 83.7 & 76.2 & 76.6 & 73.7 & 72.4 & 73.0 & 72.1 & 68.1 & 68.4 & 72.0 & 68.2 & 71.5 & 64.5 & 58.0 & 62.4 & 71.3 \\
                \citet{lample2019cross}  & Wiki & 1 & 100 & 83.2 & 76.7 & 77.7 & 74.0 & 72.7 & 74.1 & 72.7 & 68.7 & 68.6 & 72.9 & 68.9 & 72.5 & 65.6 & 58.2 & 62.4 & 70.7 \\
                \bf XLM-R\textsubscript{Base}  & CC & 1 & 100 & 85.8 & 79.7 & 80.7 & 78.7 & 77.5 & 79.6 & 78.1 & 74.2 & 73.8 & 76.5 & 74.6 & 76.7 & 72.4 & 66.5 & 68.3 & 76.2 \\
                \bf XLM-R & CC & 1 & 100 & \bf 89.1 & \bf 84.1 & \bf 85.1 & \bf 83.9 & \bf 82.9 & \bf 84.0 & \bf 81.2 & \bf 79.6 & \bf 79.8 & \bf 80.8 & \bf 78.1 & \bf 80.2 & \bf 76.9 & \bf 73.9 & \bf 73.8 & \bf 80.9 \\
                \midrule
                \multicolumn{19}{l}{\it Translate everything to English and use English-only model (TRANSLATE-TEST)} \\
                \midrule
                BERT-en & Wiki & 1 & 1 & 88.8 & 81.4 & 82.3 & 80.1 & 80.3 & 80.9 & 76.2 & 76.0 & 75.4 & 72.0 & 71.9 & 75.6 & 70.0 & 65.8 & 65.8 & 76.2 \\
                RoBERTa & Wiki+CC & 1 & 1 & \underline{\bf 91.3} & 82.9 & 84.3 & 81.2 & 81.7 & 83.1 & 78.3 & 76.8 & 76.6 & 74.2 & 74.1 & 77.5 & 70.9 & 66.7 & 66.8 & 77.8 \\
                \midrule
                \multicolumn{19}{l}{\it Fine-tune multilingual model on each training set (TRANSLATE-TRAIN)} \\
                \midrule 
                \citet{lample2019cross}  & Wiki & N & 100 & 82.9 & 77.6 & 77.9 & 77.9 & 77.1 & 75.7 & 75.5 & 72.6 & 71.2 & 75.8 & 73.1 & 76.2 & 70.4 & 66.5  & 62.4 & 74.2 \\
                \midrule
                \multicolumn{19}{l}{\it Fine-tune multilingual model on all training sets (TRANSLATE-TRAIN-ALL)} \\
                \midrule
                \citet{lample2019cross}$^{\dagger}$ & Wiki+MT & 1 & 15 & 85.0 & 80.8 & 81.3 & 80.3 & 79.1 & 80.9 & 78.3 & 75.6 & 77.6 & 78.5 & 76.0 & 79.5 & 72.9 & 72.8 & 68.5 & 77.8 \\
                \citet{huang2019unicoder} & Wiki+MT & 1 & 15 & 85.6 & 81.1 & 82.3 & 80.9 & 79.5 & 81.4 & 79.7 & 76.8 & 78.2 & 77.9 & 77.1 & 80.5 & 73.4 & 73.8 & 69.6 & 78.5 \\
                \citet{lample2019cross}  & Wiki & 1 & 100 & 84.5 & 80.1 & 81.3 & 79.3 & 78.6 & 79.4 & 77.5 & 75.2 & 75.6 & 78.3 & 75.7 & 78.3 & 72.1 & 69.2 & 67.7 & 76.9 \\
                \bf XLM-R\textsubscript{Base} & CC & 1 & 100 & 85.4 & 81.4 & 82.2 & 80.3 & 80.4 & 81.3 & 79.7 & 78.6 & 77.3 & 79.7 & 77.9 & 80.2 & 76.1 & 73.1 & 73.0 & 79.1 \\
                \bf XLM-R & CC & 1 & 100 & \bf 89.1 & \underline{\bf 85.1} & \underline{\bf 86.6} & \underline{\bf 85.7} & \underline{\bf 85.3} & \underline{\bf 85.9} & \underline{\bf 83.5} & \underline{\bf 83.2} & \underline{\bf 83.1} & \underline{\bf 83.7} & \underline{\bf 81.5} & \underline{\bf 83.7} & \underline{\bf 81.6} & \underline{\bf 78.0} & \underline{\bf 78.1} & \underline{\bf 83.6} \\
                \bottomrule
            \end{tabular}
            }
            \caption{\textbf{Results on cross-lingual classification.} We report the accuracy on each of the 15 XNLI languages and the average accuracy. We specify the dataset D used for pretraining, the number of models \#M the approach requires and the number of languages \#lg the model handles. Our \xlmr results are averaged over five different seeds. We show that using the translate-train-all approach which leverages training sets from multiple languages, \xlmr obtains a new state of the art on XNLI of $83.6$\% average accuracy. Results with $^{\dagger}$ are from \citet{huang2019unicoder}. 
            \label{tab:xnli}}
        \end{center}
    \end{table*}
}

\newcommand{\insertMLQA}{
\begin{table*}[h!]
    \begin{center}
        \resizebox{1\linewidth}{!}{
        \begin{tabular}[h]{l cc ccccccc c}
        \toprule
            {\bf Model} & {\bf train} & {\bf \#lgs} & {\bf en} & {\bf es} & {\bf de} & {\bf ar} & {\bf hi} & {\bf vi} & {\bf zh} & {\bf Avg} \\
            \midrule
            BERT-Large$^{\dagger}$ & en & 1 & 80.2 / 67.4 & - & - & - & - & - & - & - \\
            mBERT$^{\dagger}$ & en & 102 & 77.7 / 65.2 & 64.3 / 46.6 & 57.9 / 44.3 & 45.7 / 29.8 & 43.8 / 29.7 & 57.1 / 38.6 & 57.5 / 37.3 & 57.7 / 41.6 \\
            XLM-15$^{\dagger}$ & en & 15 & 74.9 / 62.4 & 68.0 / 49.8 & 62.2 / 47.6 & 54.8 / 36.3 & 48.8 / 27.3 & 61.4 / 41.8 & 61.1 / 39.6 & 61.6 / 43.5 \\
            XLM-R\textsubscript{Base} & en & 100 & 77.1 / 64.6 & 67.4 / 49.6 & 60.9 / 46.7 & 54.9 / 36.6 & 59.4 / 42.9 & 64.5 / 44.7 & 61.8 / 39.3 & 63.7 / 46.3 \\
            \bf XLM-R & en & 100 & \bf 80.6 / 67.8 & \bf 74.1 / 56.0 & \bf 68.5 / 53.6 & \bf 63.1 / 43.5 & \bf 69.2 / 51.6 & \bf 71.3 / 50.9 & \bf 68.0 / 45.4 & \bf 70.7 / 52.7  \\
            \bottomrule
        \end{tabular}
            }
            \caption{\textbf{Results on MLQA question answering} We report the F1 and EM (exact match) scores for zero-shot classification where models are fine-tuned on the English Squad dataset and evaluated on the 7 languages of MLQA. Results with $\dagger$ are taken from the original MLQA paper \citet{lewis2019mlqa}.
            \label{tab:mlqa}}
    \end{center}
\end{table*}
}

\newcommand{\insertNER}{
\begin{table}[t]
    \begin{center}
        \resizebox{1\linewidth}{!}{
        \begin{tabular}[b]{l cc cccc c}
            \toprule
            {\bf Model} & {\bf train} & {\bf \#M} & {\bf en} & {\bf nl} & {\bf es} & {\bf de} & {\bf Avg}\\
            \midrule
            \citet{lample-etal-2016-neural} & each & N & 90.74 & 81.74 & 85.75 & 78.76 & 84.25 \\
            \citet{akbik2018coling} & each & N & \bf 93.18 & 90.44 & - & \bf 88.27 & - \\
            \midrule
            \multirow{2}{*}{mBERT$^{\dagger}$} & each & N & 91.97 & 90.94 & 87.38 & 82.82 & 88.28\\
             & en & 1 & 91.97 & 77.57 & 74.96 & 69.56 & 78.52\\
             \midrule
            \multirow{3}{*}{XLM-R\textsubscript{Base}} & each & N & 92.25 & 90.39 & 87.99 & 84.60 & 88.81\\
             & en & 1 & 92.25 & 78.08 & 76.53 & 69.60 & 79.11\\
             & all & 1 & 91.08 & 89.09 & 87.28 & 83.17 & 87.66 \\
             \midrule
            \multirow{3}{*}{\bf XLM-R} & each & N & 92.92 & \bf 92.53 & \bf 89.72 & 85.81 & 90.24\\
             & en & 1 & 92.92 & 80.80 & 78.64 & 71.40 & 80.94\\
             & all & 1 & 92.00 & 91.60 & 89.52 & 84.60 & 89.43 \\
            \bottomrule
        \end{tabular}
            }
            \caption{\textbf{Results on named entity recognition} on CoNLL-2002 and CoNLL-2003 (F1 score). Results with $\dagger$ are from \citet{wu2019beto}. Note that mBERT and \xlmr do not use a linear-chain CRF, as opposed to \citet{akbik2018coling} and \citet{lample-etal-2016-neural}.
            \label{tab:ner}}
        \end{center}
       \vspace{-0.6cm}
    \end{table}
}

\newcommand{\insertAblationone}{
\begin{table*}[h!]
	\begin{minipage}[t]{0.3\linewidth}
	\begin{center}
          \includegraphics{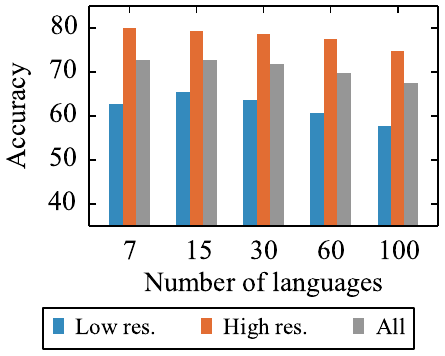}
        \captionof{figure}{The transfer-interference trade-off: Low-resource languages benefit from scaling to more languages, until dilution (interference) kicks in and degrades overall performance.}
            \label{fig:transfer_dilution}
        \vspace{-0.2cm}
	\end{center}
    \end{minipage}
    \hfill
	\begin{minipage}[t]{0.3\linewidth}
	\begin{center}
          \includegraphics{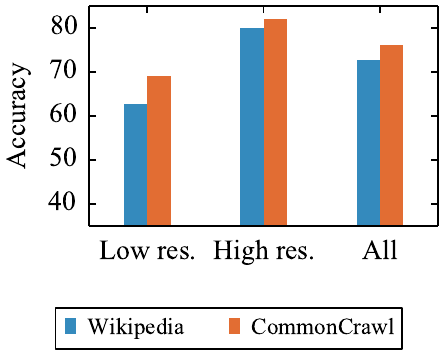}
        \captionof{figure}{Wikipedia versus CommonCrawl: An XLM-7 obtains significantly better performance when trained on CC, in particular on low-resource languages.}
            \label{fig:curse}
	\end{center}
    \end{minipage}
    \hfill
	\begin{minipage}[t]{0.3\linewidth}
	\begin{center}
          \includegraphics{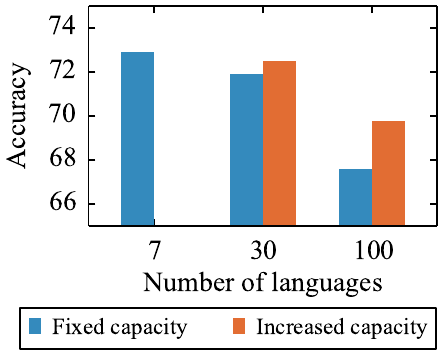}
        \captionof{figure}{Adding more capacity to the model alleviates the curse of multilinguality, but remains an issue for models of moderate size.}
            \label{fig:capacity}
	\end{center}
    \end{minipage}
    \vspace{-0.2cm}
\end{table*}
}

\newcommand{\insertAblationtwo}{
\begin{table*}[h!]
	\begin{minipage}[t]{0.3\linewidth}
        \begin{center}
          \includegraphics{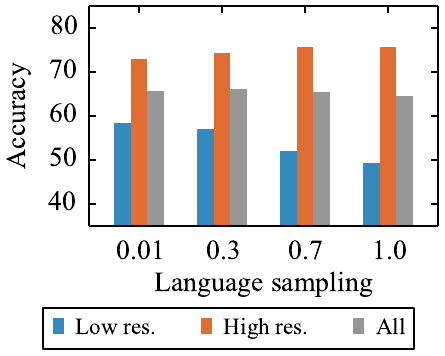}
        \captionof{figure}{On the high-resource versus low-resource trade-off: impact of batch language sampling for XLM-100.
        \label{fig:alpha}}
        \end{center}
    \end{minipage}
    \hfill
	\begin{minipage}[t]{0.3\linewidth}
        \begin{center}
          \includegraphics{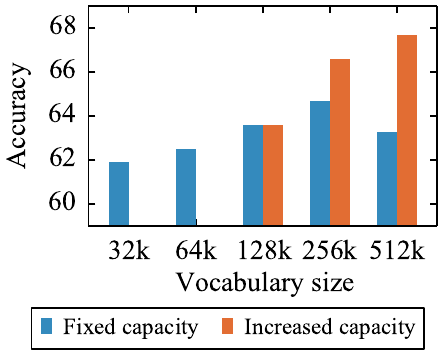}
        \captionof{figure}{On the impact of vocabulary size at fixed capacity and with increasing capacity for XLM-100.
        \label{fig:vocab}}
        \end{center}
    \end{minipage}
    \hfill
	\begin{minipage}[t]{0.3\linewidth}
        \begin{center}
          \includegraphics{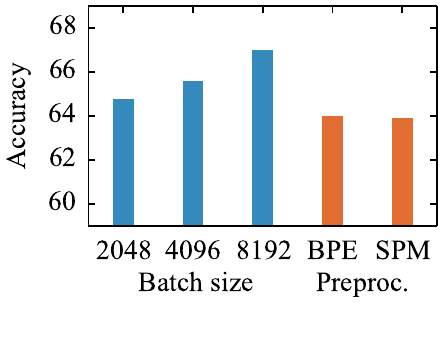}
        \captionof{figure}{On the impact of large-scale training, and preprocessing simplification from BPE with tokenization to SPM on raw text data.
        \label{fig:batch}}
        \end{center}
    \end{minipage}
        \vspace{-0.2cm}
\end{table*}
}

\newcommand{\insertMultiMono}{
    \begin{table}[h!]
        \begin{center}
            \resizebox{1\linewidth}{!}{
            \begin{tabular}[b]{l cc ccccccc c}
            \toprule
                {\bf Model} & {\bf D } & {\bf \#vocab} & {\bf en} & {\bf fr} & {\bf de} & {\bf ru} & {\bf zh} & {\bf sw} & {\bf ur} & {\bf Avg}\\
                \midrule
                \multicolumn{11}{l}{\it Monolingual baselines}\\
                \midrule
                \multirow{2}{*}{BERT} & Wiki & 40k & 84.5 & 78.6 & 80.0 & 75.5 & 77.7 & 60.1 & 57.3 & 73.4 \\
                 & CC & 40k & 86.7 & 81.2 & 81.2 & 78.2 & 79.5 & 70.8 & 65.1 & 77.5 \\
                \midrule
                \multicolumn{11}{l}{\it Multilingual models (cross-lingual transfer)}\\
                \midrule
                \multirow{2}{*}{XLM-7} & Wiki & 150k & 82.3 & 76.8 & 74.7 & 72.5 & 73.1 & 60.8 & 62.3 & 71.8 \\
                 & CC & 150k & 85.7 & 78.6 & 79.5 & 76.4 & 74.8 & 71.2 & 66.9 & 76.2 \\
                \midrule
                \multicolumn{11}{l}{\it  Multilingual models (translate-train-all)}\\
                \midrule
                \multirow{2}{*}{XLM-7} & Wiki & 150k & 84.6 & 80.1 & 80.2 & 75.7 & 78 & 68.7 & 66.7 & 76.3 \\
                 & CC & 150k & \bf 87.2 & \bf 82.5 & \bf 82.9 & \bf 79.7 & \bf 80.4 & \bf 75.7 & \bf 71.5 & \bf 80.0 \\ 
                \bottomrule
            \end{tabular}
            }
            \caption{\textbf{Multilingual versus monolingual models (BERT-BASE).} We compare the performance of monolingual models (BERT) versus multilingual models (XLM) on seven languages, using a BERT-BASE architecture. We choose a vocabulary size of 40k and 150k for monolingual and multilingual models.
            \label{tab:multimono}}
        \end{center}
       \vspace{-0.4cm}
    \end{table}
}

\newcommand{\insertGlue}{
    \begin{table}[h!]
        \begin{center}
            \resizebox{1\linewidth}{!}{
            \begin{tabular}[b]{l|c|cccccc|c}
            \toprule
                {\bf Model} & {\bf \#lgs} & {\bf MNLI-m/mm} & {\bf QNLI} & {\bf QQP} & {\bf SST} & {\bf MRPC} & {\bf STS-B} & {\bf Avg}\\
                \midrule
                BERT\textsubscript{Large}$^{\dagger}$ & 1 & 86.6/- & 92.3 & 91.3 & 93.2 & 88.0 & 90.0 & 90.2 \\
                XLNet\textsubscript{Large}$^{\dagger}$ & 1 & 89.8/- & 93.9 & 91.8 & 95.6 & 89.2 & 91.8 & 92.0 \\
                RoBERTa$^{\dagger}$ & 1 & 90.2/90.2 & 94.7 & 92.2 & 96.4 & 90.9 & 92.4 & 92.8 \\
                XLM-R & 100 & 88.9/89.0 & 93.8 & 92.3 & 95.0 & 89.5 & 91.2 & 91.8 \\
                \bottomrule
            \end{tabular}
            }
            \caption{\textbf{GLUE dev results.} Results with $^{\dagger}$ are from \citet{roberta2019}.  We compare the performance of \xlmr to BERT\textsubscript{Large}, XLNet and RoBERTa on the English GLUE benchmark.
            \label{tab:glue}}
        \end{center}
       \vspace{-0.4cm}
    \end{table}
}

\newcommand{\insertWikivsCC}{
    \begin{table*}[h]
        \begin{center}
          \includegraphics{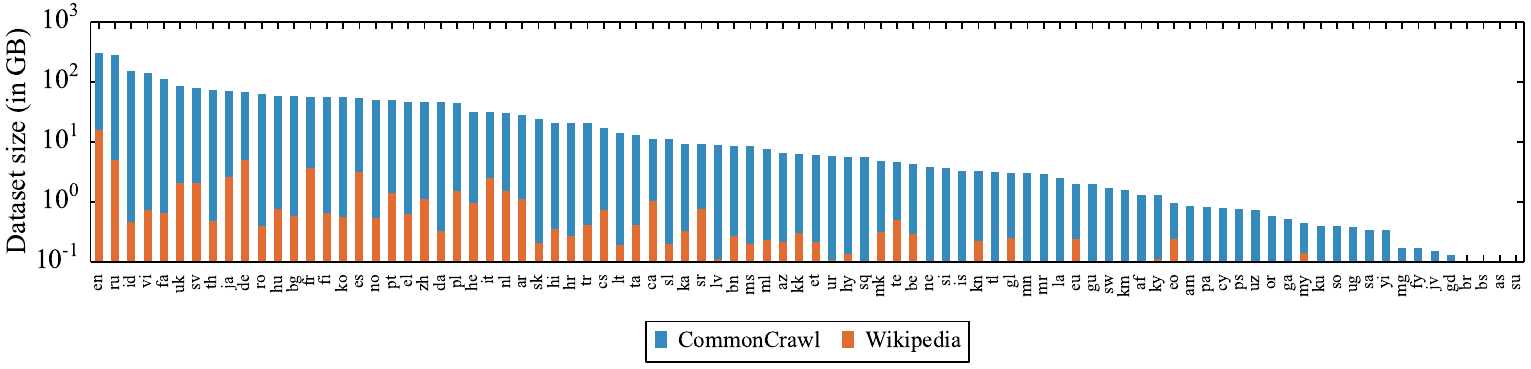}
        \captionof{figure}{Amount of data in GiB (log-scale) for the 88 languages that appear in both the Wiki-100 corpus used for mBERT and XLM-100, and the CC-100 used for XLM-R. CC-100 increases the amount of data by several orders of magnitude, in particular for low-resource languages.
        \label{fig:wikivscc}}
        \end{center}
    \end{table*}
}

\newcommand{\insertDataStatistics}{
\begin{table}[h!]
\begin{center}
\small
\begin{tabular}[b]{clrrclrr}
\toprule
\textbf{ISO code} & \textbf{Language} & \textbf{Tokens} (M) & \textbf{Size} (GiB) & \textbf{ISO code} & \textbf{Language} & \textbf{Tokens} (M) & \textbf{Size} (GiB)\\
\cmidrule(r){1-4}\cmidrule(l){5-8}
{\bf af }& Afrikaans & 242 & 1.3 &{\bf  lo }& Lao & 17 & 0.6 \\
{\bf am }& Amharic & 68 & 0.8 &{\bf  lt }& Lithuanian & 1835 & 13.7 \\
{\bf ar }& Arabic & 2869 & 28.0 &{\bf  lv }& Latvian & 1198 & 8.8 \\
{\bf as }& Assamese & 5 & 0.1 &{\bf  mg }& Malagasy & 25 & 0.2 \\
{\bf az }& Azerbaijani & 783 & 6.5 &{\bf  mk }& Macedonian & 449 & 4.8 \\
{\bf be }& Belarusian & 362 & 4.3 &{\bf  ml }& Malayalam & 313 & 7.6 \\
{\bf bg }& Bulgarian & 5487 & 57.5 &{\bf  mn }& Mongolian & 248 & 3.0 \\
{\bf bn }& Bengali & 525 & 8.4 &{\bf  mr }& Marathi & 175 & 2.8 \\
{\bf - }& Bengali Romanized & 77 & 0.5 &{\bf  ms }& Malay & 1318 & 8.5 \\
{\bf br }& Breton & 16 & 0.1 &{\bf  my }& Burmese & 15 & 0.4 \\
{\bf bs }& Bosnian & 14 & 0.1 &{\bf  my }& Burmese & 56 & 1.6 \\
{\bf ca }& Catalan & 1752 & 10.1 &{\bf  ne }& Nepali & 237 & 3.8 \\
{\bf cs }& Czech & 2498 & 16.3 &{\bf  nl }& Dutch & 5025 & 29.3 \\
{\bf cy }& Welsh & 141 & 0.8 &{\bf  no }& Norwegian & 8494 & 49.0 \\
{\bf da }& Danish & 7823 & 45.6 &{\bf  om }& Oromo & 8 & 0.1 \\
{\bf de }& German & 10297 & 66.6 &{\bf  or }& Oriya & 36 & 0.6 \\
{\bf el }& Greek & 4285 & 46.9 &{\bf  pa }& Punjabi & 68 & 0.8 \\
{\bf en }& English & 55608 & 300.8 &{\bf  pl }& Polish & 6490 & 44.6 \\
{\bf eo }& Esperanto & 157 & 0.9 &{\bf  ps }& Pashto & 96 & 0.7 \\
{\bf es }& Spanish & 9374 & 53.3 &{\bf  pt }& Portuguese & 8405 & 49.1 \\
{\bf et }& Estonian & 843 & 6.1 &{\bf  ro }& Romanian & 10354 & 61.4 \\
{\bf eu }& Basque & 270 & 2.0 &{\bf  ru }& Russian & 23408 & 278.0 \\
{\bf fa }& Persian & 13259 & 111.6 &{\bf  sa }& Sanskrit & 17 & 0.3 \\
{\bf fi }& Finnish & 6730 & 54.3 &{\bf  sd }& Sindhi & 50 & 0.4 \\
{\bf fr }& French & 9780 & 56.8 &{\bf  si }& Sinhala & 243 & 3.6 \\
{\bf fy }& Western Frisian & 29 & 0.2 &{\bf  sk }& Slovak & 3525 & 23.2 \\
{\bf ga }& Irish & 86 & 0.5 &{\bf  sl }& Slovenian & 1669 & 10.3 \\
{\bf gd }& Scottish Gaelic & 21 & 0.1 &{\bf  so }& Somali & 62 & 0.4 \\
{\bf gl }& Galician & 495 & 2.9 &{\bf  sq }& Albanian & 918 & 5.4 \\
{\bf gu }& Gujarati & 140 & 1.9 &{\bf  sr }& Serbian & 843 & 9.1 \\
{\bf ha }& Hausa & 56 & 0.3 &{\bf  su }& Sundanese & 10 & 0.1 \\
{\bf he }& Hebrew & 3399 & 31.6 &{\bf  sv }& Swedish & 77.8 & 12.1 \\
{\bf hi }& Hindi & 1715 & 20.2 &{\bf  sw }& Swahili & 275 & 1.6 \\
{\bf - }& Hindi Romanized & 88 & 0.5 &{\bf  ta }& Tamil & 595 & 12.2 \\
{\bf hr }& Croatian & 3297 & 20.5 &{\bf  - }& Tamil Romanized & 36 & 0.3 \\
{\bf hu }& Hungarian & 7807 & 58.4 &{\bf  te }& Telugu & 249 & 4.7 \\
{\bf hy }& Armenian & 421 & 5.5 &{\bf  - }& Telugu Romanized & 39 & 0.3 \\
{\bf id }& Indonesian & 22704 & 148.3 &{\bf  th }& Thai & 1834 & 71.7 \\
{\bf is }& Icelandic & 505 & 3.2 &{\bf  tl }& Filipino & 556 & 3.1 \\
{\bf it }& Italian & 4983 & 30.2 &{\bf  tr }& Turkish & 2736 & 20.9 \\
{\bf ja }& Japanese & 530 & 69.3 &{\bf  ug }& Uyghur & 27 & 0.4 \\
{\bf jv }& Javanese & 24 & 0.2 &{\bf  uk }& Ukrainian & 6.5 & 84.6 \\
{\bf ka }& Georgian & 469 & 9.1 &{\bf  ur }& Urdu & 730 & 5.7 \\
{\bf kk }& Kazakh & 476 & 6.4 &{\bf  - }& Urdu Romanized & 85 & 0.5 \\
{\bf km }& Khmer & 36 & 1.5 &{\bf  uz }& Uzbek & 91 & 0.7 \\
{\bf kn }& Kannada & 169 & 3.3 &{\bf  vi }& Vietnamese & 24757 & 137.3 \\
{\bf ko }& Korean & 5644 & 54.2 &{\bf  xh }& Xhosa & 13 & 0.1 \\
{\bf ku }& Kurdish (Kurmanji) & 66 & 0.4 &{\bf  yi }& Yiddish & 34 & 0.3 \\
{\bf ky }& Kyrgyz & 94 & 1.2 &{\bf  zh }& Chinese (Simplified) & 259 & 46.9 \\
{\bf la }& Latin & 390 & 2.5 &{\bf  zh }& Chinese (Traditional) & 176 & 16.6 \\

\bottomrule
\end{tabular}
\caption{\textbf{Languages and statistics of the CC-100 corpus.} We report the list of 100 languages and include the number of tokens (Millions) and the size of the data (in GiB) for each language. Note that we also include romanized variants of some non latin languages such as Bengali, Hindi, Tamil, Telugu and Urdu.\label{tab:datastats}}
\end{center}
\end{table}
}

\newcommand{\insertParameters}{
    \begin{table*}[h!]
        \begin{center}
            \begin{tabular}[b]{lrcrrrrrc}
            \toprule
                \textbf{Model} & \textbf{\#lgs} & \textbf{tokenization} & \textbf{L} & \textbf{$H_{m}$} & \textbf{$H_{ff}$} & \textbf{A} & \textbf{V} & \textbf{\#params}\\
                \cmidrule(r){1-1}
                \cmidrule(lr){2-3}
                \cmidrule(lr){4-8}
                \cmidrule(l){9-9}
                BERT\textsubscript{Base} & 1 & WordPiece & 12 & 768 & 3072 & 12 & 30k & 110M \\
                BERT\textsubscript{Large} & 1 & WordPiece & 24 & 1024 & 4096 & 16 & 30k & 335M \\
                mBERT & 104 & WordPiece & 12 & 768 & 3072 & 12 & 110k & 172M \\
                RoBERTa\textsubscript{Base} & 1 & bBPE & 12 & 768 & 3072 & 8 & 50k & 125M \\
                RoBERTa & 1 & bBPE & 24 & 1024 & 4096 & 16 & 50k & 355M \\
                XLM-15 & 15 & BPE & 12 & 1024 & 4096 & 8 & 95k & 250M \\
                XLM-17 & 17 & BPE & 16 & 1280 & 5120 & 16 & 200k & 570M \\
                XLM-100 & 100 & BPE & 16 & 1280 & 5120 & 16 & 200k & 570M \\
                Unicoder & 15 & BPE & 12 & 1024 & 4096 & 8 & 95k & 250M \\
                \xlmr\textsubscript{Base} & 100 & SPM & 12 & 768 & 3072 & 12 & 250k & 270M \\
                \xlmr & 100 & SPM & 24 & 1024 & 4096 & 16 & 250k & 550M \\
                GPT2 & 1 & bBPE & 48 & 1600 & 6400 & 32 & 50k & 1.5B \\
                wide-mmNMT & 103 & SPM & 12 & 2048 & 16384 & 32 & 64k & 3B \\
                deep-mmNMT & 103 & SPM & 24 & 1024 & 16384 & 32 & 64k & 3B \\
                T5-3B & 1 & WordPiece & 24 & 1024 & 16384 & 32 & 32k & 3B \\
                T5-11B & 1 & WordPiece & 24 & 1024 & 65536 & 32 & 32k & 11B \\
                \bottomrule
            \end{tabular}
            \caption{\textbf{Details on model sizes.}
            We show the tokenization used by each Transformer model, the number of layers L, the number of hidden states of the model $H_{m}$, the dimension of the feed-forward layer $H_{ff}$, the number of attention heads A, the size of the vocabulary V and the total number of parameters \#params. 
            For Transformer encoders, the number of parameters can be approximated by $4LH_m^2 + 2LH_m H_{ff} + VH_m$.
            GPT2 numbers are from \citet{radford2019language}, mm-NMT models are from the work of \citet{arivazhagan2019massively} on massively multilingual neural machine translation (mmNMT), and T5 numbers are from \citet{raffel2019exploring}. While \xlmr is among the largest models partly due to its large embedding layer, it has a similar number of parameters than XLM-100, and remains significantly smaller that recently introduced Transformer models for multilingual MT and transfer learning. While this table gives more hindsight on the difference of capacity of each model, note it does not highlight other critical differences between the models. 
            \label{tab:parameters}}
        \end{center}
       \vspace{-0.4cm}
    \end{table*}
}

%% file: xlmr.bbl
\begin{thebibliography}{40}
\expandafter\ifx\csname natexlab\endcsname\relax\def\natexlab#1{#1}\fi

\bibitem[{Akbik et~al.(2018)Akbik, Blythe, and Vollgraf}]{akbik2018coling}
Alan Akbik, Duncan Blythe, and Roland Vollgraf. 2018.
\newblock Contextual string embeddings for sequence labeling.
\newblock In \emph{COLING}, pages 1638--1649.

\bibitem[{Arivazhagan et~al.(2019)Arivazhagan, Bapna, Firat, Lepikhin, Johnson,
  Krikun, Chen, Cao, Foster, Cherry et~al.}]{arivazhagan2019massively}
Naveen Arivazhagan, Ankur Bapna, Orhan Firat, Dmitry Lepikhin, Melvin Johnson,
  Maxim Krikun, Mia~Xu Chen, Yuan Cao, George Foster, Colin Cherry, et~al.
  2019.
\newblock Massively multilingual neural machine translation in the wild:
  Findings and challenges.
\newblock \emph{arXiv preprint arXiv:1907.05019}.

\bibitem[{Bowman et~al.(2015)Bowman, Angeli, Potts, and
  Manning}]{bowman2015large}
Samuel~R. Bowman, Gabor Angeli, Christopher Potts, and Christopher~D. Manning.
  2015.
\newblock A large annotated corpus for learning natural language inference.
\newblock In \emph{EMNLP}.

\bibitem[{Conneau et~al.(2018)Conneau, Rinott, Lample, Williams, Bowman,
  Schwenk, and Stoyanov}]{conneau2018xnli}
Alexis Conneau, Ruty Rinott, Guillaume Lample, Adina Williams, Samuel~R.
  Bowman, Holger Schwenk, and Veselin Stoyanov. 2018.
\newblock Xnli: Evaluating cross-lingual sentence representations.
\newblock In \emph{EMNLP}. Association for Computational Linguistics.

\bibitem[{Devlin et~al.(2018)Devlin, Chang, Lee, and
  Toutanova}]{devlin2018bert}
Jacob Devlin, Ming-Wei Chang, Kenton Lee, and Kristina Toutanova. 2018.
\newblock Bert: Pre-training of deep bidirectional transformers for language
  understanding.
\newblock \emph{NAACL}.

\bibitem[{Grave et~al.(2018)Grave, Bojanowski, Gupta, Joulin, and
  Mikolov}]{grave2018learning}
Edouard Grave, Piotr Bojanowski, Prakhar Gupta, Armand Joulin, and Tomas
  Mikolov. 2018.
\newblock Learning word vectors for 157 languages.
\newblock In \emph{LREC}.

\bibitem[{Huang et~al.(2019)Huang, Liang, Duan, Gong, Shou, Jiang, and
  Zhou}]{huang2019unicoder}
Haoyang Huang, Yaobo Liang, Nan Duan, Ming Gong, Linjun Shou, Daxin Jiang, and
  Ming Zhou. 2019.
\newblock Unicoder: A universal language encoder by pre-training with multiple
  cross-lingual tasks.
\newblock \emph{ACL}.

\bibitem[{Johnson et~al.(2017)Johnson, Schuster, Le, Krikun, Wu, Chen, Thorat,
  Vi{\'e}gas, Wattenberg, Corrado et~al.}]{johnson2017google}
Melvin Johnson, Mike Schuster, Quoc~V Le, Maxim Krikun, Yonghui Wu, Zhifeng
  Chen, Nikhil Thorat, Fernanda Vi{\'e}gas, Martin Wattenberg, Greg Corrado,
  et~al. 2017.
\newblock Google’s multilingual neural machine translation system: Enabling
  zero-shot translation.
\newblock \emph{TACL}, 5:339--351.

\bibitem[{Joulin et~al.(2017)Joulin, Grave, and Mikolov}]{joulin2017bag}
Armand Joulin, Edouard Grave, and Piotr Bojanowski~Tomas Mikolov. 2017.
\newblock Bag of tricks for efficient text classification.
\newblock \emph{EACL 2017}, page 427.

\bibitem[{Jozefowicz et~al.(2016)Jozefowicz, Vinyals, Schuster, Shazeer, and
  Wu}]{jozefowicz2016exploring}
Rafal Jozefowicz, Oriol Vinyals, Mike Schuster, Noam Shazeer, and Yonghui Wu.
  2016.
\newblock Exploring the limits of language modeling.
\newblock \emph{arXiv preprint arXiv:1602.02410}.

\bibitem[{Kudo(2018)}]{kudo2018subword}
Taku Kudo. 2018.
\newblock Subword regularization: Improving neural network translation models
  with multiple subword candidates.
\newblock In \emph{ACL}, pages 66--75.

\bibitem[{Kudo and Richardson(2018)}]{kudo2018sentencepiece}
Taku Kudo and John Richardson. 2018.
\newblock Sentencepiece: A simple and language independent subword tokenizer
  and detokenizer for neural text processing.
\newblock \emph{EMNLP}.

\bibitem[{Lample et~al.(2016)Lample, Ballesteros, Subramanian, Kawakami, and
  Dyer}]{lample-etal-2016-neural}
Guillaume Lample, Miguel Ballesteros, Sandeep Subramanian, Kazuya Kawakami, and
  Chris Dyer. 2016.
\newblock \href {https://doi.org/10.18653/v1/N16-1030} {Neural architectures
  for named entity recognition}.
\newblock In \emph{NAACL}, pages 260--270, San Diego, California. Association
  for Computational Linguistics.

\bibitem[{Lample and Conneau(2019)}]{lample2019cross}
Guillaume Lample and Alexis Conneau. 2019.
\newblock Cross-lingual language model pretraining.
\newblock \emph{NeurIPS}.

\bibitem[{Lewis et~al.(2019)Lewis, O\u{g}uz, Rinott, Riedel, and
  Schwenk}]{lewis2019mlqa}
Patrick Lewis, Barlas O\u{g}uz, Ruty Rinott, Sebastian Riedel, and Holger
  Schwenk. 2019.
\newblock Mlqa: Evaluating cross-lingual extractive question answering.
\newblock \emph{arXiv preprint arXiv:1910.07475}.

\bibitem[{Liu et~al.(2019)Liu, Ott, Goyal, Du, Joshi, Chen, Levy, Lewis,
  Zettlemoyer, and Stoyanov}]{roberta2019}
Yinhan Liu, Myle Ott, Naman Goyal, Jingfei Du, Mandar Joshi, Danqi Chen, Omer
  Levy, Mike Lewis, Luke Zettlemoyer, and Veselin Stoyanov. 2019.
\newblock Roberta: {A} robustly optimized {BERT} pretraining approach.
\newblock \emph{arXiv preprint arXiv:1907.11692}.

\bibitem[{Mikolov et~al.(2013{\natexlab{a}})Mikolov, Le, and
  Sutskever}]{mikolov2013exploiting}
Tomas Mikolov, Quoc~V Le, and Ilya Sutskever. 2013{\natexlab{a}}.
\newblock Exploiting similarities among languages for machine translation.
\newblock \emph{arXiv preprint arXiv:1309.4168}.

\bibitem[{Mikolov et~al.(2013{\natexlab{b}})Mikolov, Sutskever, Chen, Corrado,
  and Dean}]{mikolov2013distributed}
Tomas Mikolov, Ilya Sutskever, Kai Chen, Greg~S Corrado, and Jeff Dean.
  2013{\natexlab{b}}.
\newblock Distributed representations of words and phrases and their
  compositionality.
\newblock In \emph{NIPS}, pages 3111--3119.

\bibitem[{Pennington et~al.(2014)Pennington, Socher, and
  Manning}]{pennington2014glove}
Jeffrey Pennington, Richard Socher, and Christopher~D. Manning. 2014.
\newblock \href {http://www.aclweb.org/anthology/D14-1162} {Glove: Global
  vectors for word representation}.
\newblock In \emph{EMNLP}, pages 1532--1543.

\bibitem[{Peters et~al.(2018)Peters, Neumann, Iyyer, Gardner, Clark, Lee, and
  Zettlemoyer}]{peters2018deep}
Matthew~E Peters, Mark Neumann, Mohit Iyyer, Matt Gardner, Christopher Clark,
  Kenton Lee, and Luke Zettlemoyer. 2018.
\newblock Deep contextualized word representations.
\newblock \emph{NAACL}.

\bibitem[{Pires et~al.(2019)Pires, Schlinger, and Garrette}]{Pires2019HowMI}
Telmo Pires, Eva Schlinger, and Dan Garrette. 2019.
\newblock How multilingual is multilingual bert?
\newblock In \emph{ACL}.

\bibitem[{Radford et~al.(2018)Radford, Narasimhan, Salimans, and
  Sutskever}]{radford2018improving}
Alec Radford, Karthik Narasimhan, Tim Salimans, and Ilya Sutskever. 2018.
\newblock \href
  {https://s3-us-west-2.amazonaws.com/openai-assets/research-covers/language-unsupervised/language_understanding_paper.pdf}
  {Improving language understanding by generative pre-training}.
\newblock \emph{URL
  https://s3-us-west-2.amazonaws.com/openai-assets/research-covers/language-unsupervised/language\_understanding\_paper.pdf}.

\bibitem[{Radford et~al.(2019)Radford, Wu, Child, Luan, Amodei, and
  Sutskever}]{radford2019language}
Alec Radford, Jeffrey Wu, Rewon Child, David Luan, Dario Amodei, and Ilya
  Sutskever. 2019.
\newblock Language models are unsupervised multitask learners.
\newblock \emph{OpenAI Blog}, 1(8).

\bibitem[{Raffel et~al.(2019)Raffel, Shazeer, Roberts, Lee, Narang, Matena,
  Zhou, Li, and Liu}]{raffel2019exploring}
Colin Raffel, Noam Shazeer, Adam Roberts, Katherine Lee, Sharan Narang, Michael
  Matena, Yanqi Zhou, Wei Li, and Peter~J. Liu. 2019.
\newblock Exploring the limits of transfer learning with a unified text-to-text
  transformer.
\newblock \emph{arXiv preprint arXiv:1910.10683}.

\bibitem[{Rajpurkar et~al.(2018)Rajpurkar, Jia, and Liang}]{rajpurkar2018know}
Pranav Rajpurkar, Robin Jia, and Percy Liang. 2018.
\newblock Know what you don't know: Unanswerable questions for squad.
\newblock \emph{ACL}.

\bibitem[{Rajpurkar et~al.(2016)Rajpurkar, Zhang, Lopyrev, and
  Liang}]{rajpurkar-etal-2016-squad}
Pranav Rajpurkar, Jian Zhang, Konstantin Lopyrev, and Percy Liang. 2016.
\newblock \href {https://doi.org/10.18653/v1/D16-1264} {{SQ}u{AD}: 100,000+
  questions for machine comprehension of text}.
\newblock In \emph{EMNLP}, pages 2383--2392, Austin, Texas. Association for
  Computational Linguistics.

\bibitem[{Sang(2002)}]{sang2002introduction}
Erik~F Sang. 2002.
\newblock Introduction to the conll-2002 shared task: Language-independent
  named entity recognition.
\newblock \emph{CoNLL}.

\bibitem[{Schuster et~al.(2019)Schuster, Ram, Barzilay, and
  Globerson}]{schuster2019cross}
Tal Schuster, Ori Ram, Regina Barzilay, and Amir Globerson. 2019.
\newblock Cross-lingual alignment of contextual word embeddings, with
  applications to zero-shot dependency parsing.
\newblock \emph{NAACL}.

\bibitem[{Siddhant et~al.(2019)Siddhant, Johnson, Tsai, Arivazhagan, Riesa,
  Bapna, Firat, and Raman}]{siddhant2019evaluating}
Aditya Siddhant, Melvin Johnson, Henry Tsai, Naveen Arivazhagan, Jason Riesa,
  Ankur Bapna, Orhan Firat, and Karthik Raman. 2019.
\newblock Evaluating the cross-lingual effectiveness of massively multilingual
  neural machine translation.
\newblock \emph{AAAI}.

\bibitem[{Singh et~al.(2019)Singh, McCann, Keskar, Xiong, and
  Socher}]{singh2019xlda}
Jasdeep Singh, Bryan McCann, Nitish~Shirish Keskar, Caiming Xiong, and Richard
  Socher. 2019.
\newblock Xlda: Cross-lingual data augmentation for natural language inference
  and question answering.
\newblock \emph{arXiv preprint arXiv:1905.11471}.

\bibitem[{Socher et~al.(2013)Socher, Perelygin, Wu, Chuang, Manning, Ng, and
  Potts}]{socher2013recursive}
Richard Socher, Alex Perelygin, Jean Wu, Jason Chuang, Christopher~D Manning,
  Andrew Ng, and Christopher Potts. 2013.
\newblock Recursive deep models for semantic compositionality over a sentiment
  treebank.
\newblock In \emph{EMNLP}, pages 1631--1642.

\bibitem[{Tan et~al.(2019)Tan, Ren, He, Qin, Zhao, and
  Liu}]{tan2019multilingual}
Xu~Tan, Yi~Ren, Di~He, Tao Qin, Zhou Zhao, and Tie-Yan Liu. 2019.
\newblock Multilingual neural machine translation with knowledge distillation.
\newblock \emph{ICLR}.

\bibitem[{Tjong Kim~Sang and De~Meulder(2003)}]{tjong2003introduction}
Erik~F Tjong Kim~Sang and Fien De~Meulder. 2003.
\newblock Introduction to the conll-2003 shared task: language-independent
  named entity recognition.
\newblock In \emph{CoNLL}, pages 142--147. Association for Computational
  Linguistics.

\bibitem[{Vaswani et~al.(2017)Vaswani, Shazeer, Parmar, Uszkoreit, Jones,
  Gomez, Kaiser, and Polosukhin}]{transformer17}
Ashish Vaswani, Noam Shazeer, Niki Parmar, Jakob Uszkoreit, Llion Jones,
  Aidan~N. Gomez, Lukasz Kaiser, and Illia Polosukhin. 2017.
\newblock Attention is all you need.
\newblock In \emph{Advances in Neural Information Processing Systems}, pages
  6000--6010.

\bibitem[{Wang et~al.(2018)Wang, Singh, Michael, Hill, Levy, and
  Bowman}]{wang2018glue}
Alex Wang, Amapreet Singh, Julian Michael, Felix Hill, Omer Levy, and Samuel~R
  Bowman. 2018.
\newblock Glue: A multi-task benchmark and analysis platform for natural
  language understanding.
\newblock \emph{arXiv preprint arXiv:1804.07461}.

\bibitem[{Wenzek et~al.(2019)Wenzek, Lachaux, Conneau, Chaudhary, Guzman,
  Joulin, and Grave}]{wenzek2019ccnet}
Guillaume Wenzek, Marie-Anne Lachaux, Alexis Conneau, Vishrav Chaudhary,
  Francisco Guzman, Armand Joulin, and Edouard Grave. 2019.
\newblock Ccnet: Extracting high quality monolingual datasets from web crawl
  data.
\newblock \emph{arXiv preprint arXiv:1911.00359}.

\bibitem[{Williams et~al.(2017)Williams, Nangia, and
  Bowman}]{williams2017broad}
Adina Williams, Nikita Nangia, and Samuel~R Bowman. 2017.
\newblock A broad-coverage challenge corpus for sentence understanding through
  inference.
\newblock \emph{Proceedings of the 2nd Workshop on Evaluating Vector-Space
  Representations for NLP}.

\bibitem[{Wu et~al.(2019)Wu, Conneau, Li, Zettlemoyer, and
  Stoyanov}]{wu2019emerging}
Shijie Wu, Alexis Conneau, Haoran Li, Luke Zettlemoyer, and Veselin Stoyanov.
  2019.
\newblock Emerging cross-lingual structure in pretrained language models.
\newblock \emph{ACL}.

\bibitem[{Wu and Dredze(2019)}]{wu2019beto}
Shijie Wu and Mark Dredze. 2019.
\newblock Beto, bentz, becas: The surprising cross-lingual effectiveness of
  bert.
\newblock \emph{EMNLP}.

\bibitem[{Xie et~al.(2019)Xie, Dai, Hovy, Luong, and Le}]{xie2019unsupervised}
Qizhe Xie, Zihang Dai, Eduard Hovy, Minh-Thang Luong, and Quoc~V Le. 2019.
\newblock Unsupervised data augmentation for consistency training.
\newblock \emph{arXiv preprint arXiv:1904.12848}.

\end{thebibliography}
